\documentclass{sig-alternate}

\newlength\myindent
\newcommand\bindent{%
  \begingroup
  \setlength{\itemindent}{\myindent}
  \addtolength{\algorithmicindent}{\myindent}
}
\newcommand\eindent{\endgroup}

\begin{document}
%
% --- Author Metadata here ---
%\conferenceinfo{WOODSTOCK}{'97 El Paso, Texas USA}
%\CopyrightYear{2007} % Allows default copyright year (20XX) to be over-ridden - IF NEED BE.
%\crdata{0-12345-67-8/90/01}  % Allows default copyright data (0-89791-88-6/97/05) to be over-ridden - IF NEED BE.
% --- End of Author Metadata ---

\newcommand{\ie}{\textit{i.e.}}
\newcommand{\eg}{\textit{e.g.}}

\title{Privacy Preserving QoE Modeling using Collaborative Learning}
%\subtitle{[Extended Abstract]
%\titlenote{A full version of this paper is available as
%\textit{Author's Guide to Preparing ACM SIG Proceedings Using
%\LaTeX$2_\epsilon$\ and BibTeX} at

\numberofauthors{3} %  in this sample file, there are a *total*
\author{
% 1st. author
\alignauthor
Selim Ickin\\
       \affaddr{Ericsson Research}\\
       \email{selim.ickin\\@ericsson.com}
% 2nd. author
\alignauthor
Konstantinos Vandikas\\
       \affaddr{Ericsson Research}\\
       \email{konstantinos.vandikas\\@ericsson.com}
% 3rd. author
\alignauthor 
Markus Fiedler\\
       \affaddr{Blekinge Inst. of Tech.}\\
		\email{markus.fiedler\\@bth.se}
}
% There's nothing stopping you putting the seventh, eighth, etc.
% author on the opening page (as the 'third row') but we ask,
% for aesthetic reasons that you place these 'additional authors'
% in the \additional authors block, viz.
%\additionalauthors{Additional authors: John Smith (The Th{\o}rv{\"a}ld Group,
%email: {\texttt{jsmith@affiliation.org}}) and Julius P.~Kumquat
%(The Kumquat Consortium, email: {\texttt{jpkumquat@consortium.net}}).}
% Just remember to make sure that the TOTAL number of authors
% is the number that will appear on the first page PLUS the
% number that will appear in the \additionalauthors section.

\maketitle
\begin{abstract}
Machine Learning based Quality of Experience\,(QoE) models potentially suffer from over-fitting due to limitations including low data volume, and limited participant profiles. This prevents models from becoming generic. Consequently, these trained models may under-perform when tested outside the experimented population. One reason for the limited datasets, which we refer in this paper as small QoE data lakes, is due to the fact that often these datasets potentially contain user sensitive information and are only collected throughout expensive user studies with special user consent. Thus, sharing of datasets amongst researchers is often not allowed. In recent years, privacy preserving machine learning models have become important and so have techniques that enable model training without sharing datasets but instead relying on secure communication protocols. Following this trend, in this paper, we present \emph{Round-Robin} based  Collaborative Machine Learning model training, where the model is trained in a sequential manner amongst the collaborated partner nodes. We benchmark this work using our customized \emph{Federated Learning} mechanism as well as conventional \emph{Centralized} and \emph{Isolated Learning} methods.
\end{abstract}

%A category including the fourth, optional field follows...
%\category{D.2.8}{Software Engineering}{Metrics}[complexity measures, performance measures]



\section{Introduction}
Today, QoE models are developed based on isolated data lakes within the premises of researchers, as sharing of data is often not preferred or allowed. As such, data privacy is preserved but the models might have the risk of being not sufficiently representative. There is an increasing trend that the data sets collected via QoE experiments are becoming semi-public; only accessible by special request\,\cite{qualinet-database}. Thus, many similar models can be obtained from the same data set with different settings. Hence there is a need for collaboration techniques for internal communication in-between researchers regarding the details of problem formulation, and model development. Within the premise of this work, we consider a model to be preserving privacy if it can be trained without the need for moving user-sensitive raw data in between researchers. Instead of raw data transfer, in order to sustain privacy, the obtained weights from the trained neural networks are transferred. This is due to the fact that there are techniques in the state of the art (\cite{bonawitz}, \cite{dwork}) that can be applied to different machine learning model exchange techniques thus ensuring privacy.

In this paper, we present a collaborative learning mechanism, to the best of our knowledge for the first time in the area of QoE modeling, where every researcher at individual data lakes contributes to the final model by partially training the model on their individual isolated data sets. Then, the trained model weights are shared in between via various collaborative learning techniques comprising \emph{Round-Robin learning} (RRL) and \emph{Federated Learning} (FL). In RRL, the training process happens at an isolated node at a time and the trained model is shared with other models to continue on training with the other existing datasets in a round robin manner, also known as \emph{ring all reduce}\,\cite{ringallreduce}. In FL, iteratively,  every independent model at isolated nodes trains using Stochastic Gradient Descent (SGD) on the existing isolated datasets, shares the learned weights via a central node (which we refer as master node) and then eventually aggregates the weights received from all nodes followed by an iterative weight broadcasting procedure. We perform experiments in order to compare the collaborative learning QoE model accuracy values with Isolated Learning (IL) QoE  models.

This paper is structured as follows. We begin with presenting the related studies within the scope of the paper in Section\,\ref{sec-relatedwork}. We describe the study of four machine learning training scenarios in Section\,\ref{sec-mltrainingmechs}. In the Collaborative Learning technique, we present how accuracy can be improved, where we mainly focus on the Neural Network (NN) algorithm. The corresponding dataset and  model details are given in Section\,\ref{sec-datasetmodelling}. For the IL technique, we start with benchmarking  results from a popular and simpler machine learning model, Decision Tree (DT), which is often used in QoE modeling due to its advantages in interpretability. The findings are collectively presented with the discussion on the results in Section\,\ref{sec-results}. The paper is concluded in Section\, \ref{sec-conclusion}.

\section{Related Work}\label{sec-relatedwork}
ML algorithms such as Decision Trees, Random Forests are a few of the most commonly used techniques in the QoE literature\,  \cite{mlmodel-survey-qoe}. Support Vector Machines (SVM) have been used earlier in QoE Modeling as they often perform well in small datasets \cite{svm-qoe-youtube}. These models are hard to use for Collaborative Learning as a continuation of training after a model transfer is a challenge. Especially, when larger datasets are used, then there are better alternatives such as Neural Networks, whose weights can be updated using  Collaborative Learning techniques.
\begin{figure*}[htbp!]
\centering
{\includegraphics[width=0.85\linewidth]{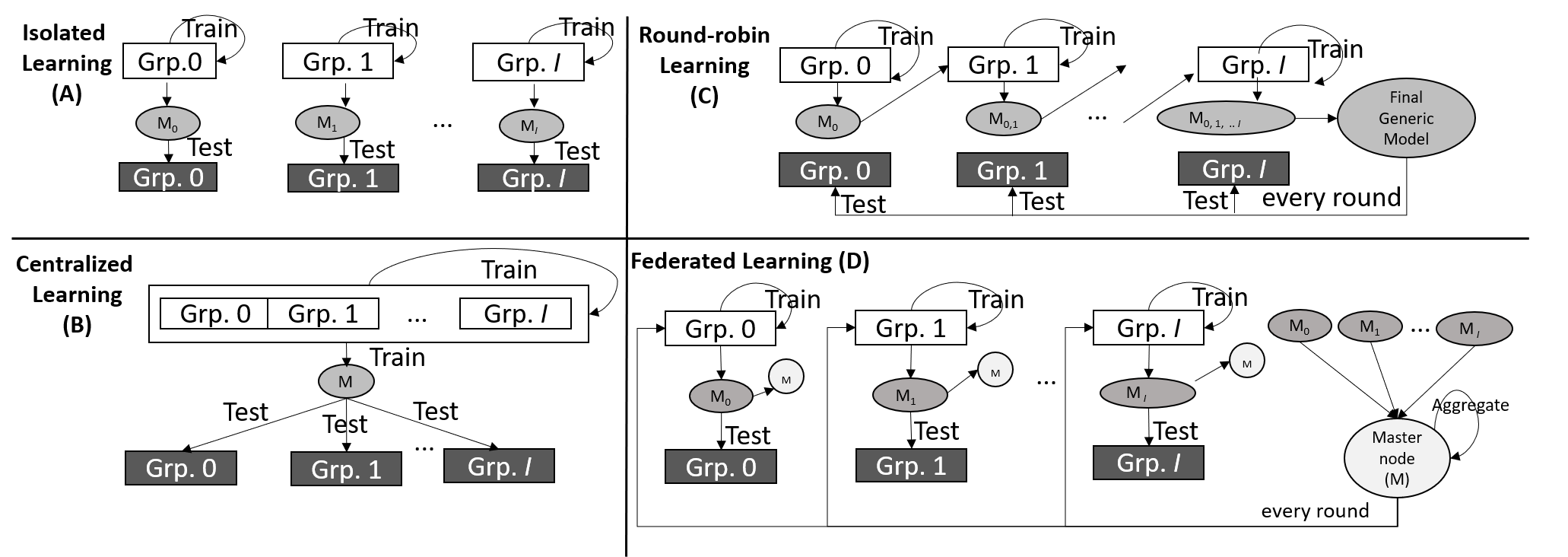}}
\caption{Comparison of different learning techniques.}
\label{schemafig}
\end{figure*}
\section{ML Model Training MECHANISMS}\label{sec-mltrainingmechs}
\begin{table}[htbp]
\small
\caption{Scenario comparison summary.}
\begin{center}
\begin{tabular}{|c|c|c|c|}
\hline
%\textbf{Table}&\multicolumn{3}{|c|}{\textbf{Table Column Head}} \\
%\cline{2-4}
\textbf{} & \textbf{\textit{Centralized}}& \textbf{\textit{Isolated}}& \textbf{\textit{Collaborative}} \\
\hline
\textit{Data transfer} &High&None& None  \\
\hline
\textit{Weight transfer} & None & None & Low  \\
\hline
\textit{Privacy preserving} & No&Yes&Yes  \\
\hline
\textit{Training} & Master & Workers & Workers  \\
\hline
\end{tabular}
\label{scenariosummary}
\end{center}
\end{table}
The experiments are performed in four scenarios: Centralized (CL), Isolated (IL), Round-Robin Learning (RRL), and Federated Learning (FL). The scenarios are described as follows. The latter two scenarios are collaborative learning which we propose mainly in this paper, where the first two are studied for bench-marking purposes. The benefit of using Collaborative Learning is summarized in Table\,\ref{scenariosummary}, with reduced network utilization, while preserving data privacy.

\subsection{Centralized Learning (CL)} In this conventional scenario, the datasets from all worker nodes are transferred to one node, and then model training is performed on this node as given in Fig.\ref{schemafig}.B. The product of this process is a singular model which can be made available using a model serving infrastructure. If user data is sensitive, or if data cannot be moved due to legislative reasons, this way of training may not be preferred.

\subsection{Isolated Learning (IL)} No data is transferred from the worker nodes, and all worker nodes train on their individual datasets as depicted Fig.\ref{schemafig}.A. The model accuracy is limited to the data in the isolated nodes. If a worker node has sufficient data, a representable model can be obtained but it will never benefit from other worker nodes that may collect new data. We compare Decision Tree (DT) and Neural Network (NN) algorithms.

\subsection{Round Robin Learning (RRL)} The worker nodes train on their individual datasets, with size $n$, without sharing any data with each other. Instead, they share neural weights, $w$, in a sequential manner as described with   Fig.~\ref{schemafig}.C. Worker $i$ starts training on its own isolated local dataset, then sends the trained model to the next worker $i+1$ and continues on training. This processes continues until the full round is complete\,\cite{ringallreduce}. After one round is completed, the final model is evaluated on the individual worker's testsets. This way, by collaboratively training on the model, a model can be trained without sharing data that consists of input features, $x$, and target variable, $y$. The pseudo code for this training processes is given in Alg.\,\ref{alg-rr}.

\begin{algorithm}
\caption{Round-robin Learning with ascending order.}
\label{alg-rr}
\begin{algorithmic}
\REQUIRE $n \geq 0 \vee x \neq null \vee y\neq null$
\STATE initialize $W$, \STATE $i \leftarrow 0$

\STATE $w^i \leftarrow W$
\FORALL {round $r \in R$}
\bindent
    \STATE $i \leftarrow 0$
    \WHILE {$i < I-1$}
        \bindent
            %\STATE $w_i^r \leftarrow SGD(w_i)$
            %\STATE $w_i^{(r+1)} \leftarrow w_i^r$
            \STATE$w_{i+1}^{r}  \leftarrow SGD(x_i^{r},w_i^{r};y_i^{r})$
            \STATE increment $i$
        \eindent
    \ENDWHILE
\eindent
\ENDFOR
\end{algorithmic}
\end{algorithm}

\subsection{Federated Learning (FL)}
In Federated Learning we trade transfer of data to the master node with moving neural networks to the worker nodes. The Federated Learning mechanism is sketched in Fig.\,\ref{schemafig}.D, and it works as follows. First an initial weight matrix, $W$, is initialized by a master node, and then it is broadcasted to the workers. Next, the local workers train on their own localized datasets with some learning rate (lr), \ie, the step size of updating weights in every epoch, and then send their trained neural weights back to the master node. The master node, once it has received the anticipated amount of neural weights, $K$, for averaging, performs the federated averaging and sends the averaged weights, $\overline{w^r}$ of round, $r$, back to the workers. This cyclic training process continues until the model accuracy reaches a saturation point, $R_\textrm{saturation}$.
The pseudo code for this training processes is given in Alg.\,\ref{alg-fl}. In order to estimate the health of the federation, workers are allowed to record the accuracy achieved after each training session and for the final evaluation only after the model reached some maturity. This is the round where the accuracy of the models reaches an approximate steady state value.

\begin{algorithm}
\caption{Federated Learning}
\label{alg-fl}
\begin{algorithmic}
\REQUIRE $n \geq 0 \vee x \neq null \vee y\neq null$
\STATE initialize $W$
\FORALLP {worker $i \in I$}
\bindent
    \STATE $w^i \leftarrow W$
\eindent
\ENDFOR
\FORALL {round $r \in R$}
\bindent
    \FORALLP {worker $i \in I$ }
    \bindent
        \STATE $w_i^r \leftarrow SGD(x_i^{r},w_i^{r};y_i^{r})$
        \STATE $weightDict[i] \leftarrow w_i^{r}$
    \eindent
    \ENDFOR
    \IF {$len(weightDict) > K$}
    \bindent
        \STATE $\overline{w^r} \leftarrow \mathrm{average}(weightDict)$
        \FORALLP {worker $i \in I$ }
        \bindent
            \STATE $w_i^{r+1} \leftarrow \overline{w^r}$
        \eindent
        \ENDFOR
        \STATE $weightDict \leftarrow null$
    \eindent
    \ENDIF
\eindent
\ENDFOR
\end{algorithmic}
\end{algorithm}

\subsection{Collaborative Learning Prototype}
For the purpose of performing RRL and FL based evaluation for this experiment, we designed and implemented a prototype that implements both protocols. As far as the Federated Learning protocol is considered we rely on the description provided in \cite{mcmahan16}. We chose to develop our own prototype instead of using a framework such as Tensorflow Federated\,\cite{tff} (TFF) since at the moment when this experiment took place such a framework was not available. More so, Tensorflow Federated is meant to target Android devices while in our case we are more interested in performing Federated Learning between different compute nodes in a cloud environment. However, since the purpose of this prototype is to implement protocols for both RRL and FL, we rely on Tensorflow (TF) to train a neural network. In that way we avoid re-inventing best practise for hardware accelerated neural network training and inference while at the same time retain the option of switching Tensorflow with other implementations such as Pytorch\,\cite{pytorch} or Ray\,\cite{ray18}.

\begin{figure}[htbp!]
{\includegraphics[width=\linewidth]{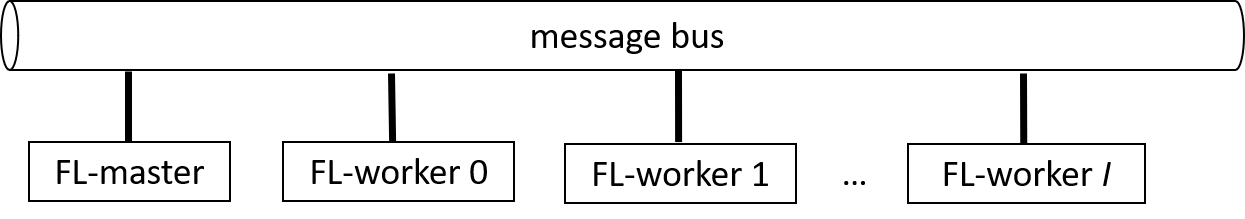}}
\caption{Basic architecture of our prototype for Federated Learning}
\label{proto_arch}
\end{figure}

To share weights between the different workers we rely on a message bus since each training is a separate process that runs on a different computer. A high-level overview of the prototype is illustrated in Fig.\,\ref{proto_arch}. When it comes to RRL we follow the same protocol as FL  but instead of sending neural weights to the master node, we send the neural weights to the next worker using the message queue. In both cases and to maintain a small network footprint, we rely on \emph{pyarrow}\footnote{ \url{https://pypi.org/project/pyarrow/}} serialization/deserialization technique before we place our payload in the message queue to be transferred to the next worker node.

The model training time for FL and RRL is compared. RRL is a sequential training, \ie, training on one node starts after training is completed on another node, hence the total training time is the sum of individual training time on all workers. On the other hand, in FL training, the training takes place in parallel; all workers train on their individual datasets simultaneously, and share weights after each round. Shared weights, stored in a weight dictionary, \emph{weightDict}, are averaged and then sent back to the workers. Therefore the total training time is the sum of training time at every round. The estimated training time of the two collaborative learning mechanisms, RRL and FL is given in Eqns.\,\ref{RR-time-eq} and \ref{FL-time-eq}, respectively.
\begin{equation}
\label{RR-time-eq}
T_\textrm{RRL, training} = \sum_{i=0}^{I} T_i
\end{equation}

\begin{equation}
\label{FL-time-eq}
T_\textrm{FL,training}= \max(T_r) : r=R_\textrm{saturation}
\end{equation}
In FL, $R_\textrm{saturation}$ is the minimum round id that the performance has saturated, and $T_r$ is the time it takes for the worker to complete round, $r$. In RRL, $T_i$ is the time it takes for worker $i$ to complete training.
\section{Dataset and Modeling}\label{sec-datasetmodelling}
\subsection{Dataset}
\begin{figure}[htbp!]
{\includegraphics[width=\linewidth]{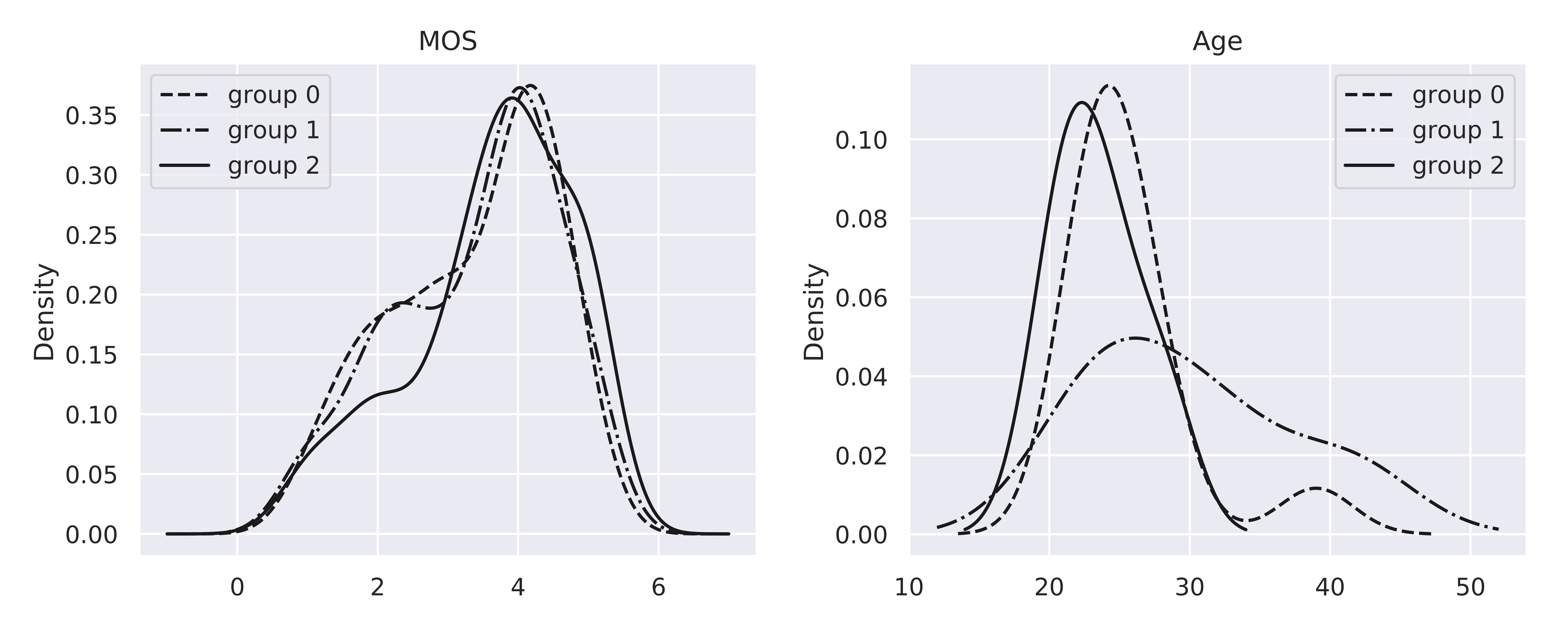}}
\caption{Kernel Density Estimate (KDE) plots for the MOS ratings and the age of the users on each group. Reasonable bandwidths are chosen in the visualisation to avoid under- or over-fitting.}
\label{age_mos_kde}
\end{figure}

The public web QoE dataset which is available at \cite{data} is used in the experiments. The dataset is artificially and arbitrarily divided in three different groups, where the assumption is that these three isolated groups are located at different data centers and are not allowed to share raw data amongst each other. The users in the dataset are grouped with respect to their user ids. User ids below 22 belong to group (Grp.) 0, user ids less than 37 belong to Grp. 1, and the remaining users are grouped in Grp. 2. We intended to have balanced data size on all user groups while setting  the thresholds. The dataset contains in total 32 users, where there are 13, 10, and 9 users in groups 0, 1, and 2, respectively. The user opinion scores are collected using a 9 step scale from 1 to 5 with step size 0.5. The probability distribution of the age of the users and the MOS scores given by the users in the three groups are given in Fig.\,\ref{age_mos_kde}. The age distribution on Grp. 1 seems significantly different as compared to user Grp. 0 and Grp. 2. In overall, user groups 0 and 1 are rather similar in recorded MOS scores, while Grp. 2 comprises of rather higher MOS scores.

Mean Opinion Score (MOS), 5 level, is often used in the literature, however there are also examples where binary classifiers (``low'' and ``high'' using 3.5 MOS scores as threshold) are preferred in the modeling \cite{qoe-binary-classification}. Since the dataset consists of only 292 samples, we transformed the opinion score values of the users into 2 classes such that the scores that are below 3.5 are considered to be poor, while the scores that are higher or equal to 3.5 are classified as being in the good class. The input features used in the modeling are listed as follows: i) maximum \emph{downlink bandwidth (dl. bw.)} set in the network emulator; ii) browsing time until the rating prompt which is the \emph{surfing duration (dur. surfing)};   and iii) time consumed during the rating process, \emph{prompt duration (dur. prompt)}. The features such as round trip time delay and uplink  bandwidth did not have entropy on the modeling, thus are removed from the feature set.  \par
The descriptive statistics on the dataset and the model are given in Tables\,\ref{datasummary} and \ref{modelconfig}. The users in Grp. 2, has tendency to give better QoE ratings, while the  Grp. 0 users register the lowest QoE ratings. Still, the distribution of the three datasets are not very different than each other.

\begin{table}[htbp]
\small

\caption{Dataset descriptive statistics.}
\begin{center}
\begin{tabular}{|c|c|c|c|c|c|c|}
\hline
\textbf{feature} & \textbf{Grp.}& \textbf{mean}& \textbf{std} & \textbf{max} & \textbf{min}& \textbf{size}  \\
\hline\hline
dl. bw\,(kbit/s)&0&376.0&347.9&1024&32&160\\
\hline
dl. bw\,(kbit/s)&1&361.4&351.7&1024&32&136\\
\hline
dl. bw\,(kbit/s)&2&365.1&353.7&1024&32&122\\
\hline
\hline
dur. surfing\,(s)&0&161.2&19.4&247&3&160\\
\hline
dur. surfing\,(s)&1&165.0&12.9&214&138&136\\
\hline
dur. surfing\,(s)&2&158.0&11.5&214&126&122\\
\hline
\hline
dur. prompt\,(s)&0&15.8&7.2&49&6&160\\
\hline
dur. prompt\,(s)&1&15.8&9.9&77&6&136\\
\hline
dur. prompt\,(s)&2&16.4&9.5&73&5&122\\
\hline
\hline
Binary MOS &0&0.5&0.49&1&0&160\\
\hline
Binary MOS &1&0.6&0.48&1&0&136\\
\hline
Binary MOS &2&0.7&0.45&1&0&122\\
\hline\hline
\end{tabular}
\label{datasummary}
\end{center}
\end{table}

\begin{table}[htbp]
\small

\caption{Experimented data and model config.}
\begin{center}
\begin{tabular}{|c|c|}
\hline
User  & Grp. 0 ($uid<22$)\\
group &  Grp. 1 ($22<=uid<37$)\\
dataset & Grp. 2 ($37<uid$)\\
\hline
Train/test ratio& 60\,\% trainset, 40\,\% testset\\
\hline
Input features& dl bw, dur. surfing, dur. prompt \\
%&duration prompt\\
\hline
Target variable &  Binary MOS (1 if $>=3.5$, else $=0$) \\
\hline
Decision tree& max tree: 2, 3, 5. Criterion: Gini\\
\hline
Neural Net.& 1 hidden layer (4, 8, 16, 32 neurons) \\
& epochs: 50, 100, 200, 400, 800 \\
& learning rate: 0.001, Dropout: $30\,\%$\\
& l2 norm: 0.02, Activation: ReLu, Softmax.\\
&early stop(val. loss, patience: 10 epochs)\\
\hline
\end{tabular}
\label{modelconfig}
\end{center}
\end{table}

\subsection{ML Models}
The Decision Tree (DT) models are hierarchical binary trees, comprising root and leaf nodes, that partition the dataset based on information gain. Gini criterion is used on the nodes, such that the class label distribution of the classes are taken into consideration in decisions. We experimented with three different DT model configurations where the maximum depth of the trees are 2, 3, and 5, respectively.\par
The Neural Network (NN) model is evaluated with different numbers of hidden layers and neurons such as 1 hidden layer with 4, 8, 16, 32 neurons with different epochs comprising 50, 100, 200, 400, 800. Since the dataset size is small, we do not go beyond 1 hidden layer, but vary the hidden layer size instead. Different randomization on the initial weights often causes different local minimums, and this is known to vanish with deeper neural networks. That is one of the reasons why deep neural networks are often preferred on large datasets. We executed over 100 independent experiments for each finding to reduce confidence intervals of  the results.
% TODO you mention that  different neural networks are used but in the next section
The model is a 3 layer network with a single hidden layer. ReLu activation function is used in the hidden layer. L2 regularization with a factor of 0.02 is used to avoid over-fitting, and a Dropout with $30\%$, and Batch Normalization is introduced before the final fully connected layer. Adam optimizer with a categorical cross entropy for optimization evaluation, and also the learning rate is set to 0.001. The number of epochs in all experiments is 400. We set an early stop criteria in training, such that if there is no reduction in the validation loss for 10 epochs, it stops the training earlier than 400 epochs. Summary of the tested model parameters is given in Table\,\ref{modelconfig}. 
\subsection{Model Evaluation}\label{sec-modelevaluation}
We use ROC (Receiver Operating Characteristics) \emph{AUC (Area Under the Curve)} score for model performance evaluation, which is commonly used for binary classification problems in the literature. For the goal of having a good model, good estimated separation between a poor or good QoE, it is of high desire to maximize ROC AUC score. The model, in the inference phase, outputs a probability for the corresponding classes, and based on the cutoff threshold, the model decides the samples that are predicted to be good if the probability is above the cutoff threshold, otherwise they are estimated to be at the poor class. ROC AUC curve is computed with multiple cutoff decimal points between 0 and 1, and then plot, for each cutoff probability threshold, the corresponding True Positive Rate (\% of true poor class predictions to the total actual poor class samples) with respect to the False Positive Rate (\% of false poor class predictions to the total actual good class samples), by comparing the estimated classes to the actual ones. At the end of the probability scan, the area under the curve yields the AUC score, which is aimed to be close to 1.0.

\section{Results}\label{sec-results}

\subsection{Isolated Learning (IL)}\label{subsec:isolated}
Two machine learning algorithms, one simple DT and one rather more complex NN, are studied with different hyper parameters to model QoE. The experiments are performed to understand and find out the best hyper parameters. We let the isolated models to train at best effort, \ie, tuned the hyper parameters until the AUC  did not improve anymore, and then used the model parameters for the final evaluation. We evaluated the performance of the isolated models both within the same user group and also across groups. 
\begin{table}[htbp]
\small

\caption{Mean AUC with $95\,\%$ Confidence Interval\,(CI) of the group models through 100 independent experiments, where all workers perform isolated trained (DT).}
\begin{center}
\begin{tabular}{|c|c|c|c|}
\hline
\textbf{Train On}& \textbf{Test On} & \textbf{Test On}  & \textbf{Test On}  \\
DT(maxdepth) & Grp. 0& Grp. 1 &Grp. 2 \\
 & AUC & AUC &AUC  \\
\hline
\textbf{\emph{DT (2)}} Grp. 0 & 0.69(0.01) & 0.74 (0.01) &0.73(0.01) \\
\hline
\textbf{\emph{DT (2)}} Grp. 1 &  0.67(0.01) & 0.74(0.01) &0.69(0.01) \\
\hline
\textbf{\emph{DT (2)}} Grp. 2 &0.69(0.01)  &0.74(0.01) & 0.72(0.01) \\
\hline\hline

\emph{DT (3)} Grp. 0 & 0.67(0.01) & 0.73 (0.01) &0.71(0.01) \\
\hline
\emph{DT (3)} Grp. 1 & 0.66(0.01) & 0.73(0.01)&0.68(0.01) \\
\hline
\emph{DT (3)} Grp. 2 &0.67(0.01)  &0.72(0.01) & 0.71(0.01) \\
\hline\hline

\emph{DT (5)} Grp. 0 & 0.66(0.01) & 0.70 (0.01) &0.73(0.01) \\
\hline
\emph{DT (5)} Grp. 1 & 0.64(0.01) & 0.68(0.01)&0.66(0.01) \\
\hline
\emph{DT (5)} Grp. 2 &0.64(0.01)  &0.69(0.01) & 0.61(0.01) \\
\hline\hline
\end{tabular}
\label{overallsummaryresultsDT}
\end{center}
\end{table}
Maximum tree depth size of 2, amongst the three tested tree sizes, is observed to be performing decent with mean AUC of 0.71, as given in Table\,\ref{overallsummaryresultsDT}. As given in Table\,\ref{overallsummaryresultsILNN}, NN(16,400) model (with 16 neurons and 400 epochs) performed the best amongst the tested number of neurons and epochs, with a mean AUC of 0.75, when trained and tested on the same group, hence we use this neural network architecture and settings in CL, RRL, and FL.\par 
The trained model on one group is tested on other groups for benchmarking purposes and to study how a transfer of one model in one user group performs on another user group. In our experiments, we developed the models to minimize the under- over-fitting issues, hence the results presented on Isolated Learning scenario show that the models trained on one node is representative, hence performs with similar accuracies when tested on other user groups. Therefore, our aim here is not to show that isolated models over-fit, but rather to show that the accuracy of the models can further be improved using collaborative learning without sharing raw data in between.

 \begin{table}[htbp]
 \small

\caption{Mean AUC (with $95\,\%$ CI) of the group models through 100 independent experiments, where all workers perform isolated trained (NN).}
\begin{center}
\begin{tabular}{|c|c|c|c|}
\hline
\textbf{Train On}& \textbf{Test On} & \textbf{Test On}  & \textbf{Test On}  \\
NN(Neurons,  & Grp. 0& Grp. 1 &Grp. 2 \\
 epochs)& AUC & AUC &AUC  \\
 \hline
 \hline
 \emph{NN(4, 50)} Grp. 0 & 0.63(0.03) & 0.64(0.03) &0.63(0.03) \\
\hline
\emph{NN(4, 50)} Grp. 1 & 0.60(0.03) & 0.60(0.03)& 0.63(0.03) \\
\hline
\emph{NN(4, 50)} Grp. 2 &0.62(0.03)  & 0.62(0.03) &  0.61(0.03) \\
\hline\hline
%\emph{NN (4, 100)} Grp. 0 & 0.66(+/- 0.03) & 0.70 (+/-0.03) &0.69(+/-0.03) \\
%\hline
%\emph{NN (4, 100)} Grp. 1 & 0.68(+/-0.02) & 0.72(+/-0.03)&0.71(+/-0.03) \\
%\hline
%\emph{NN (4, 100)} Grp. 2 & 0.62(+/-0.03)  &0.65(+/-0.04) &  0.65(+/-0.04) \\

%\hline\hline
%\emph{NN (4, 200)} Grp. 0 & 0.69(+/- 0.02) & 0.75(+/-0.02) &0.74(+/-00.02) \\
%\hline
%\emph{NN (4, 200)} Grp. 1 & 0.70(+/-0.02) & 0.75(+/-0.02)&0.73(+/-00.02) \\
%\hline
%\emph{NN (4, 200)} Grp. 2 & 0.65(+/-0.02)  &0.70(+/-0.03) &  0.69(+/-0.03) \\

%\hline\hline
%\emph{NN (8, 50)} Grp. 0 & 0.64(+/-0.01) & 0.67 (+/-0.01) &0.67(+/-0.01) \\
%\hline
%\emph{NN (8, 50)} Grp. 1 &  0.66(+/-0.01) & 0.69(+/-0.01)&0.68(+/-0.01) \\
%\hline
%\emph{NN (8, 50)} Grp. 2 &0.64(+/-0.01)  &0.67(+/-0.01) & 0.66(+/-0.01) \\
\hline\hline
\emph{NN(8, 100)} Grp. 0 & 0.73(0.01) & 0.73 (0.01) &0.73(0.01) \\
\hline
\emph{NN(8, 100)} Grp. 1 & 0.74(0.01) & 0.73(0.01)&0.73(0.01) \\
\hline
\emph{NN(8, 100)} Grp. 2 & 0.71(0.01)  &0.72(0.01) & 0.71(0.01) \\
\hline\hline
%\emph{NN (8, 200)} Grp. 0 & 0.71(+/-0.02) & 0.77(+/-0.02) & 0.75(+/-0.02) \\
%\hline
%\emph{NN (8, 200)} Grp. 1 & 0.72(+/-0.02) & 0.77(+/-0.02)&0.76(+/-0.02) \\
%\hline
%\emph{NN (8, 200)} Grp. 2 &0.70(+/-0.02)  &0.75(+/-0.02) & 0.73(+/-0.02) \\
%\hline\hline

%\emph{NN (8, 400)} Grp. 0 &0.72 (+/-0.01) & 0.79(+/-0.01) & 0.78(+/-0.01) \\
%\hline
%\emph{NN (8, 400)} Grp. 1 & 0.73(+/-0.01) & 0.80(+/-0.01)&0.79(+/-0.01) \\
%\hline
%\emph{NN (8, 400)} Grp. 2 &0.70(+/-0.02)  &0.75(+/-0.02) &  0.73(+/-0.02) \\
%\hline\hline

%\emph{NN (8, 800)} Grp. 0 & 0.74(+/-0.01) &0.79(+/-0.01) &0.77(+/-0.01) \\
%\hline
%\emph{NN (8, 800)} Grp. 1 &  0.75(+/-0.01) & 0.80(+/-0.01)& 0.79(+/-0.01) \\
%\hline
%\emph{NN (8, 800)} Grp. 2 & 0.70(+/-0.02)  &0.74 (+/-0.03) & 0.72(+/-0.01) \\

%\hline\hline
\textbf{\emph{NN(16, 400)} Grp. 0} & 0.75(0.01) & 0.76(0.01) &0.74(0.01) \\
\hline
\textbf{\emph{NN(16, 400)} Grp. 1} & 0.77(0.01) & 0.76(0.01)& 0.76(0.01) \\
\hline
\textbf{\emph{NN(16, 400)} Grp. 2} & 0.74(0.01)  & 0.73(0.01) & 0.73(0.01) \\
\hline\hline

%\emph{NN (16, 800)} Grp. 0 & 0.73(+/-0.01) &0.78(+/-0.01) & 0.77(+/-0.01) \\
%\hline
%\emph{NN (16, 800)} Grp. 1 &  0.74(+/-0.01) & 0.78(+/-0.01)&  0.77(+/-0.01) \\
%\hline
%\emph{NN (16, 800)} Grp. 2 & 0.74(+/-0.01)  & 0.78(+/-0.01) &0.76 (+/-0.01) \\
%\hline\hline

%\emph{NN (32, 800)} Grp. 0 & 0.73(+/-0.01) &0.80(+/-0.01) & 0.79(+/-0.01) \\
%\hline
%\emph{NN (32, 800)} Grp. 1 &  0.75(+/-0.01) & 0.80(+/-0.01)&  0.79(+/-0.01) \\
%\hline
%\emph{NN (32, 800)} Grp. 2 & 0.73(+/-0.01)  & 0.78(+/-0.01) &0.77 (+/-0.01) \\
%\hline\hline
\end{tabular}
\label{overallsummaryresultsILNN}
\end{center}
\end{table}

\subsection{Collaborative Learning}
\begin{figure}[htbp!]
{\includegraphics[width=\linewidth]{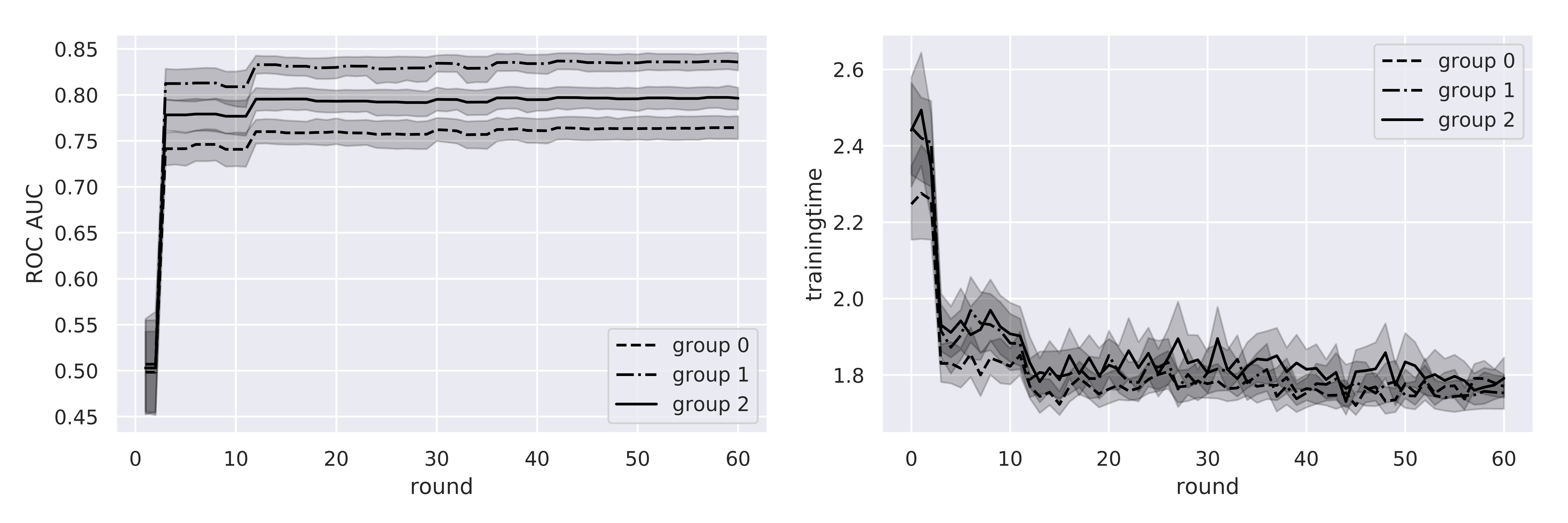}}
\caption{Mean (with $95\%$ CI) training performance(left) and time(right) over rounds in FL.}
\label{fl_perf_time}
\end{figure}

Best effort NN model and parameters (16 neurons, 400 epochs) from Section\,\ref{subsec:isolated} are chosen as we did not see any improvement in AUC in isolated scenarios with higher number of neurons. We aim to use these minimum settings to draw the bottom line AUC scores for the collaborative scenarios. The ROC AUC values from CL, RRL, and FL mechanisms are depicted for all workers (user groups) in Table\,\ref{overallsummaryresultsNN}. We present the AUC scores from the $15^\mathrm{th}$ round for the Federated Learning scenario since at the $15^\mathrm{th}$ round, all worker nodes are observed to achieve a saturation point in the ROC AUC score as given in the left figure in Fig.\,\ref{fl_perf_time}. The mean round training time also saturated after 15 rounds of training as shown in the right side of the same figure.   

%
%\begin{figure}[t!]%
%    \subfloat[Federated Learning accuracy over rounds.\label{flperf}]{{\includegraphics[width=0.5\linewidth]{figures/training_over_rounds.png} }}%
%    \subfloat[Federated Learning training time over rounds. \label{fltime}]{{\includegraphics[width=0.5\linewidth]{figures/training_over_rounds_time.png} }}%
%    \caption{Mean (with $95\%$ CI) training performance and time over rounds in FL training scenario. }%
%\end{figure}

\begin{table}[htbp]
\small

\caption{Mean AUC (with $95\,\%$ CI) of the group models through 100 independent experiments, where all workers collaborates to build one model (NN).}
\begin{center}
\begin{tabular}{|c|c|c|c|}
\hline
\textbf{Train On}& \textbf{Test On} & \textbf{Test On}  & \textbf{Test On}   \\
NN model & Grp. 0& Grp. 1 &Grp. 2 \\
NN(16,400) & AUC & AUC &AUC  \\
 &  &  &  \\
\hline
%\emph{Isolated} Group 0 &0.74(+/-0.01) & 0.80(+/-0.01) & 0.79(+/-0.01) \\
%\hline
%\emph{Isolated} Group 1 & 0.74(+/-0.01) & 0.79(+/-0.01)& 0.79(+/-0.01) \\
%\hline
%\emph{Isolated} Group 2 & 0.73(+/-0.01)  & 0.77(+/-0.01) & 0.76(+/-0.01) \\
\hline
CL &0.76(0.01) &0.83(0.01) &0.81(0.01) \\
\hline
FL& & & \\
 ($R_\textrm{saturation} = 15$) &0.75(0.01) &0.83(0.01) &0.79(0.01)\\
\hline
RRL& 0.78(0.002) & 0.78(0.001) &0.77(0.002) \\
\hline
\end{tabular}
\label{overallsummaryresultsNN}
\end{center}
\end{table}
\subsection{Comparison Between Methods}
Considering the results shown in Table~\ref{overallsummaryresultsNN}, we observe that CL yields the highest AUC (in particular for Grp. 2), which is not surprising.
However, AUC provided by FL is not significantly lower\footnote{indicated by touching/overlapping confidence intervals.}. RRL yields some lower performance but for Grp. 0, for which the AUC matches the one by FL.
All those methods yield AUC scores well above 0.77, with 0.83 upper boundary. The FL model AUC is observed to be on par with the RRL in overall. \par
On the other hand, it is difficult to see how much lower the AUC of the best isolated cases is. For instance, DT\,(2) (c.f.~Table~\ref{overallsummaryresultsDT}) yields AUC values in the order of 0.69 to 0.74, and NN\,(16, 400) yields AUC scores between 0.73 and 0.76, all of them (but FL on Grp. 0) are significantly lower than the results of the Collaborative Learning. In overall, Grp. 1 and Grp. 2 benefits from Collaborative Learning, in both of the two techniques. Grp. 0 rather benefits only with the RRL. Moreover, isolated models might even perform better for other groups than the one they have been trained for, e.g.~DT\,(2) trained on Grp. 0 yields a higher AUC for Grp. 2 than DT\,(2) trained on Grp. 2. This also indicates that the isolated models  (although performs worse as compared to CL, RRL, and FL), and might also yield sufficient representation in overall when trained on sufficient datasets. 

\begin{table}[htbp]
\small

\caption{Mean(with 95\,\% CI) training time (s) comparison from over 100 independent experiments (NN).}
\begin{center}
\begin{tabular}{|c|c|c|c|c|c|}
\hline
\textbf{IL}&\textbf{IL} &\textbf{IL}  &\textbf{CL} & \textbf{FL} & \textbf{RRL} \\
\textbf{Grp. 0}&\textbf{Grp. 1} &\textbf{Grp. 2}  & & per round  & 1 cycle \\
\hline
2.11&2.24&2.13&2.58& 2.69 &2.55\\
(0.45)&(0.43)&(0.39)&(0.08)&(0.74)&(0.28)\\
\hline
\end{tabular}
\label{overallsummaryresultsNNTime}
\end{center}
\end{table}

The total training time required for the trained models to reach a saturation is measured on all scenarios as given in Table\,\ref{overallsummaryresultsNNTime}. Isolated Learning scenarios has the least training time due to the limited size of dataset on each, as expected. Interestingly, when the models are trained in a sequential RRL fashion, the bottleneck training time turned out to be the first training data lake, which is observable as follows. The training on the first node takes around 1.9\,s, continuation of the training on Grp. 1 and Grp. 2 adds 0.3\,s each. In RRL, given that the first training node (which in this study is Grp. 0) has representative model on the other groups, which we think that this is the cause of the fact that first training process takes longer as compared to the remaining training phases on other nodes. In other words, the nodes that come after the first training does not need to train from scratch, but only does fine tuning on the pre-trained model, and stops due to the early stop configuration, hence yielding a shorter training time. Similarly in FL, initial rounds \ie, the initial training phase significantly takes longer as compared to the following rounds as given in the right figure in Fig.\,\ref{fl_perf_time}. 

Amongst the collaborative learning approaches, there are pro's and con's between FL and RRL. Within one cycle training time (approx. 2.5\,s), the RRL based model reaches a minimum AUC of $0.77$ on all nodes, which is higher than when the training on the nodes take place independently and in an isolated manner. In this scenario, each node sends the weights directly to the next node, hence the next node can reveal the previously trained node's model, this technique might not be considered as privacy protecting as compared to FL. In FL, since each node only shares the weights with the master node, and the weights are only shared after the averaging, hence revealing of the individual nodes is significantly harder. We observed that the FL training time, although comparable to RRL training per round, but in total takes longer to converge probably due to aggregation technique and round trip communication delays. 
Overall, collaborative learning mechanism outperform the models that were trained via isolated manner. Via collaborative learning, similar performances are achieved as compared to the training via centralized manner. Either of the collaborative learning technique can be suggested as alternative QoE model development.
%\begin{table}[htbp]
%\caption{Mean RR sequential training time (s) over nodes with 95\,\% CI.}
%\begin{center}
%\begin{tabular}{|c|c|c|}
%\hline
%\textbf{Grp. 0}&\textbf{Grp. 1} &\textbf{Grp. 2}  \\
%\hline
%2.86&0.68&0.73\\
%+/-0.48&+/-0.31&+/-0.28\\
%\hline
%\end{tabular}
%\label{nodesRRTime}
%\end{center}
%\end{table}

%\section{Model Interpretation}\label{sec-int}
%\emph{NB: It seems as if neither time nor space are  sufficient, and we might dig into the models. Maybe we should leave this to an extended version.}

\section{Conclusion}\label{sec-conclusion}
In this paper, we present that collaborative machine learning as potential tool that can be suggested in QoE modeling. NN model accuracy outperforms the isolated decision tree models when trained either as an isolated, or in a collaborative manner. We study Federated Learning (FL) and Round Robin Learning (RRL) to show that on par accuracy can be achieved without sharing sensitive data amongst researchers. This enables achieving on par results to centralized learning while protecting privacy issues.\par
Training in collaborative learning is more straight forward when there are existing labels, in other words, for those use cases that are applicable to supervised learning. Within the scope of QoE, this is often the case where the user labels are collected to supervise or train the machine learning models, where input features are mostly QoS metrics. Hence, collaborative supervised learning is highly suggested for QoE modeling.\par 
We are aware of the limitation that the evaluation of the proposed methods within the scope of QoE is performed with only one dataset that is publicly available. The data distribution on the user groups were not also significantly different from each other, therefore testing the collaborative learning mechanisms on other available public QoE dataset is scheduled for future work.

\bibliographystyle{abbrv}
\bibliography{main}  % sigproc.bib is the name of the Bibliography in this case

\begin{thebibliography}{10}

\bibitem{pytorch}
Pytorch.
\newblock \url{https://www.pytorch.org}, Accessed: 2019-06-04.

\bibitem{tff}
Tensorflow {F}ederated.
\newblock \url{https://www.tensorflow.org/federated/}, Accessed: 2019-06-04.

\bibitem{ringallreduce}
baidu-allreduce.
\newblock \url{https://github.com/baidu-research/baidu-allreduce/}, Accessed:
  2019-06-14.

\bibitem{data}
Web browsing {QoE} subjective test dataset {V} 1.0.
\newblock \url{https://www.schatz.cc/downloads/web-dataset/}, Accessed:
  2019-06-14.

\bibitem{qualinet-database}
Qualinet database.
\newblock \url{http://dbq.multimediatech.cz}, Accessed: 2019-06-18.

\bibitem{mlmodel-survey-qoe}
S.~Aroussi et~al.
\newblock Survey on machine learning-based {QoE}-{QoS} correlation models.
\newblock {\em International Conference on Computing, Management and
  Telecommunications (ComManTel)}, pages 200--204, 2014.

\bibitem{bonawitz}
K.~Bonawitz et~al.
\newblock Practical secure aggregation for privacy-preserving machine learning.
\newblock {\em In {\em Proceedings of the 2017 ACM SIGSAC Conference on
  Computer and Communications Security}, CCS '17, pages , New York, USA. ACM.},
  pages 1175--1191, 2017.

\bibitem{dwork}
J.~Braams and C.~Dwork.
\newblock {Differential Privacy}.
\newblock {\em Springer US}, pages 338--340, 2011.

\bibitem{mcmahan16}
H.~Brendan~McMahan et~al.
\newblock Federated learning of deep networks using model averaging.
\newblock {\em \em CoRR}, 2018.

\bibitem{svm-qoe-youtube}
T.~Ho{\ss}feld et~al.
\newblock Quantification of {Y}outube {QoE} via crowdsourcing.
\newblock In {\em 2011 IEEE International Symposium on Multimedia}, pages
  494--499, 2011.

\bibitem{ray18}
P.~Moritz et~al.
\newblock Ray: A distributed framework for emerging {AI} applications.
\newblock {\em In {\em 13th {USENIX} Symposium on Operating Systems Design and
  Implementation ({OSDI} 18)}}, pages 561--577, 2018.

\bibitem{qoe-binary-classification}
I.~Orsolic et~al.
\newblock Youtube {QoE} estimation from encrypted traffic: Comparison of test
  methodologies and machine learning based models.
\newblock {\em {QoMEX'18}}, 2018.

\end{thebibliography}

\end{document}